\documentclass[10pt,twocolumn,letterpaper]{article}

\usepackage{cvpr}              

\usepackage{graphicx}
\usepackage{amsmath}
\usepackage{amssymb}
\usepackage{booktabs}
\usepackage{enumitem}
\usepackage[dvipsnames]{xcolor}
\usepackage{float}
\usepackage{subcaption}

\usepackage[pagebackref=true,breaklinks=true,letterpaper=true,colorlinks,bookmarks=false,linkcolor=blue]{hyperref}

\usepackage[capitalize]{cleveref}
\crefname{section}{Sec.}{Secs.}
\Crefname{section}{Section}{Sections}
\Crefname{table}{Table}{Tables}
\crefname{table}{Tab.}{Tabs.}

\def\cvprPaperID{49} 
\def\confName{CVPR}
\def\confYear{2022}

\begin{document}

\title{Pretraing method for image-text transformers?} 
\title{How to pre-train image-language transformers for open-vocabulary  
tasks?} 

\title{Pre-training image-language transformers for open-vocabulary  
tasks}

\author{AJ Piergiovanni\\
Google Research\\

\and

\and
Weicheng Kuo\\
Google Research\\

\and
Anelia Angelova\\
Google Research\\

}
\maketitle

\begin{abstract}
We present a pre-training approach for vision and language transformer models, which is based on a mixture of diverse tasks.
We explore both the use of image-text captioning data in pre-training, which does not need additional supervision, as well as object-aware strategies to pre-train the model. We evaluate the method on a number of text-generative vision+language tasks, such as Visual Question Answering, visual entailment and captioning, and demonstrate large gains over standard pre-training methods.
\end{abstract}

\vspace{-5mm}
\section{Introduction}

Using vision and text transformers jointly to learn vision language interactions is an important problem, with applications ranging from captioning, retrieval, Visual Question Answering (VQA), and others. Training these models is an open area of research, where one major unknown is what pre-training mechanisms to use.  Existing works, such as CLIP or ALIGN \cite{clip,align}, use large-scale web-based datasets 
of image-text pairs, 
and apply contrastive loss during pre-training. This data is unlabeled and often noisy, and the collected text descriptions refer to  
prominent objects in the image. Using additionally labeled data, which provides valuable semantic information, has been under-explored.
For example, OpenImages \cite{openimages} contains image-level and localization (box-level) labels. 
This data provides less noisy labeling for a large variety of objects in images, and can yield cleaner signals and better data efficiency during training. 

A closely coupled question is what kind of pre-training tasks are most suitable for
image-text transformer models. 
Some approaches use versions of the contrastive loss, as mentioned above~\cite{clip, align}, which is a powerful technique but less suitable for text-generative problems such as image captioning or open-vocabulary VQA.
Other approaches borrow losses from the text domain applied to image-language, such as captioning~\cite{cc3m} or masking losses, e.g., MLM~\cite{vilbert2020}. These methods can be widely applied as they do not require extra supervision, but might be inefficient or even inadequate at extracting semantic information. 
Some of these have been extended to focus on objects, typically by using off-the-shelf object detection~\cite{vilbert2020,li2020oscar,wang2022objectaware}, which clearly limits applicability to a small set of categories.

Furthermore, for VQA tasks, most prior works focus on a classification problem, using a limited set of answers (e.g., 3000 for VQA2.0) or by a fixed set of pre-defined outputs. However, a generalized system will be able to output free-form text for any question.
Similarly, captioning tasks are open-vocabulary by definition. 
We focus on these vision-language tasks in the open-vocabulary setting, although they are more challenging. This lets us investigate the impact of different pre-training tasks for the entire model, by evaluating the generated text.

We propose a pre-training approach as a mixture of variety of pre-training tasks, on both VQA and captioning. We find that combining multiple cross-modal and language-based object-aware tasks is beneficial.
Our results show large improvements over previous pre-training methods.   

\section{Related Works}
Standard methods for image-text models use an image encoder 
to obtain vision features, then a Transformer \cite{vaswani2017attention} to process the combined image and language tokens \cite{simvlm,vilbert2020,chen2020uniter}. We use a similar architecture. Where we differ from the previous works is in the diversity of the pre-training sources. 
Rather than using large image-text data \cite{clip,align} or pre-trained object detectors \cite{vilbert2020,li2020oscar,wang2022objectaware}, we use image-text tasks generated from image or box level labels. We work with open-vocabulary text generation, which is harder.

\vspace{-3mm}
\section{Model}
In this work, we build an image-language transformer model which uses both an image and a text sequence input and outputs a generated text sequence. We use this model for a number of vision language tasks, specifically VQA, visual entailment and captioning, and, as mentioned, with free-form open-vocabulary generated text outputs.
Our model follows the standard encoder-decoder transformer style. We use convolutional layers to extract image patches~\cite{xiao2021early} and use those as the vision tokens. We use a T5 \cite{T5} model to obtain text features from the language model and to generate the output text.
The model is trained to generate text using a cross-entropy loss. Our model is derived from T5-base and has 330M parameters.

\section{Pre-training Approach}
We focus on creating pre-training tasks and using a collection of tasks in a `pre-training mixture'. The tasks are diverse in several dimensions: 1) amount of supervision/labels and accompanying noise and 2) the task the model is trying to solve. We consider two types of tasks: Cross-Modal Image-Text pre-training (CM) (Section~\ref{sec:it}) and Object-Aware (OA) (Section~\ref{sec:loc}). We utilize only the OpenImages V4 dataset~\cite{openimages} to obtain different image-text tasks for all experiments. 
To allow image-text pre-training, the Localized Narratives dataset~\cite{localizednarratves}, which provides additional caption annotations for OpenImages, is used (only its OpenImages subset). 
This allows us to keep the image data constant, while only varying the tasks.

The pre-trained model is then applied to a number of downstream tasks: VQA (VQA2.0 \cite{agrawal2015vqa}, GQA \cite{hudson2019gqa}, visual entailment (SNLI-VE \cite{snli-ve}) and captioning (MSCOCO \cite{chen2015cococaptions}). 
We use these evaluations to demonstrate the effect of pre-training across a variety of use-cases. 
In the open-vocabulary setting we address, the answer is only counted as correct if it matches {\it exactly} one of the ground truth answers. We chose to evaluate in this challenging setting as it uses the entire trained encoder-decoder model.

\subsection{Image-Text Pre-training Tasks}
\label{sec:it}
To learn image-language interactions in these tasks, we use a mixture of four tasks for pre-training: captioning, completion, image-text matching (ITM) and masked language modeling (MLM). 
We refer to these tasks as Cross-Modal (CM) Image-Text tasks, since no additional supervision is needed from the image-language data (which can be obtained by automatic processes) and they are often used as such in prior works.
For these tasks, we use the Localized Narratives~\cite{localizednarratves} annotations to OpenImages.


\subsection{Object-aware tasks}
\label{sec:loc}
We construct additional tasks using object-specific information. We use both the localization class label and image-level class label in OpenImages~\cite{openimages} to obtain object-level labels. The tasks are designed to teach the model different aspects of image-text and object data, especially in the case of non-exhaustively annotated objects. The tasks are:
\begin{enumerate}
    \item Input: `List all objects' Output: `[obj1], [obj2], ...'
    \item Input: `Does [object] exist?' Output: Yes/No
    \item Input: `Does [obj1], [obj2] and/or [obj3] exist?' \\Output: Yes/No
    \item Input: `Which of [obj1], [obj2] and [obj3] exist?' \\Output: [obj1], [obj2].
\end{enumerate}

These tasks use the model in different ways. For example, listing all the objects requires the decoder to generate the names for all the objects, however, since the image has some objects, which are not annotated, this task can harm the model. The object existence task tests the model's language encoder and ability to find the object in the image, but does not make the decoder learn object names. The multiple object existence task forces the model to learn more subtle language meanings, such as the difference between `and' and `or', and the ability to find 
a set of objects. Finally, the last task makes the model output the names of the objects present in the image, but only from those specified in the input. This is similar to the list objects task, but will not penalize the model as the first task does.

\subsection{Hard Negatives}
We further want to explore the use of negative examples, 
in both Localized Narratives (LN) and OpenImages (OI). We specifically compare the use of `hard' and `easy' negative samples. For LN, the negatives are used for the ITM task. The `easy' negatives come from sampling a random caption from the dataset. The `hard' negative examples are obtained by taking a noun in the caption and replacing it with a different, but related noun, e.g., by finding a close noun in WordNet.  
For OI, the `easy' negatives are randomly sampled class names not present in the image. OpenImages also has some human verified labels, both for objects in the image and as well as objects not in the image. The initial labels are machine generated, and some are human verified. We use as the human verified negatives as the `hard' negatives in this setting, for the object exists tasks. These negatives should thus be harder for the model to learn.

\subsection{Mixture Pre-Training}
Our pre-training approach is effectively realizing a mixture of the abovementioned image-text pre-training tasks and object-aware tasks. In our experiments below, we find that this is most advantageous (Sections~\ref{sec:results} and~\ref{sec:results_cap}). In all experiments, we equally weight each task in the mixture.

\section{Experiments}

We compare the main pre-training approach to a number of alternatives. More specifically, we experiment with both single task training and mixtures of the various tasks described above. When training on mixtures of these datasets, the number of training steps is kept fixed. That is, the model sees the same number of samples each time, though the text data associated with each image naturally varies. We evaluate on open-vocabulary VQA and image captioning tasks.

\subsection{Results on Visual Question Answering tasks}
\label{sec:results}
\begin{figure}
    \centering
    \vspace{-0.3cm}
    \includegraphics[width=\linewidth]{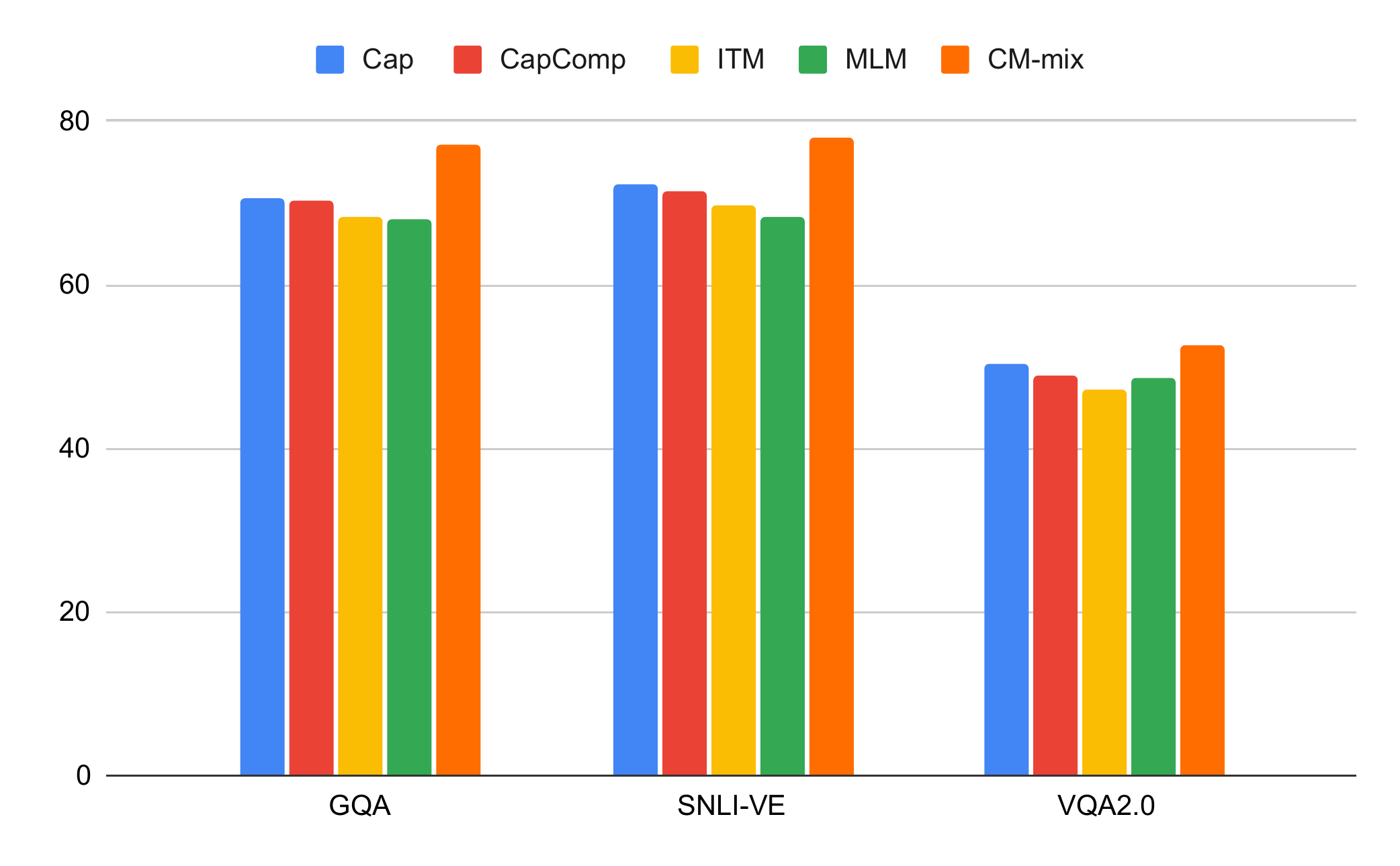}
    \caption{Comparison of the Cross-Modal Image-Language (CM) tasks and mixtures for pre-training. A mixture of all four tasks (CM-mix) performs best. 
    Multiple VQA tasks are used for downstream evaluation, all in open-vocabulary setting.}
    \label{fig:ln-pretrain}
     \vspace{-0.2cm}
\end{figure}

We first evaluate the Cross-Modal Image Language (CM) tasks, specifically evaluating individual task contributions and also the performance of the model, when they are mixed together (Fig. \ref{fig:ln-pretrain}).
We find that captioning is a strong pre-training, but there is benefit from adding the other tasks as well. A mixture of these tasks is better than any individual ones, which is consistent across a number of datasets.
%
Of note is that the mixture of tasks can give quite significant boost for some datasets. 

\begin{figure}
    \centering
    \vspace{-0.3cm}
    \includegraphics[width=\linewidth]{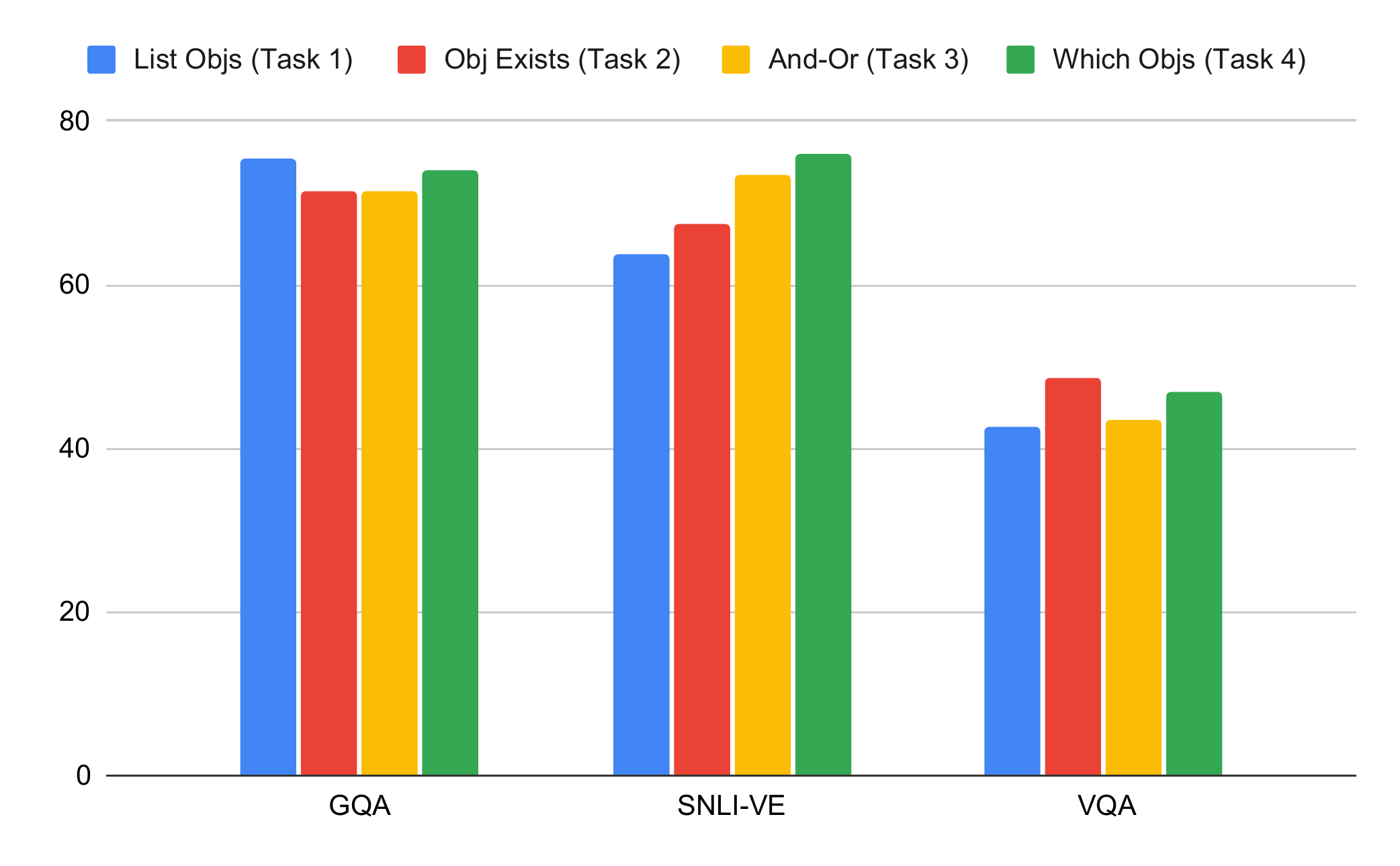}
    \caption{Performance results for single Object-Aware (OA) pre-training tasks on several VQA datasets. 
   Individual task success vary depending on the dataset.
    }
    \label{fig:single_tasks}
    \vspace{-0.2cm}
\end{figure}

Similarly, we show the results of using individual Object-Aware (OA) tasks for pre-training. Figure \ref{fig:single_tasks} shows the results. We find that for each dataset, different tasks are beneficial. Overall, the `which objects' task (task 4) is most helpful of the individual tasks. The `listing objects' task does well on GQA, but worse on the others. Each task causes the model to learn different things, which we found affects the downstream performance differently.

\begin{table*}[]
    \centering
    \begin{tabular}{l|cccc}
    \toprule
         & GQA & SNLI-VE & VQA2.0 \\
    \midrule
    Captioning Baseline  & 70.7 & 72.4 & 50.4 \\
    MLM Baseline  & 67.9 & 68.3 & 48.5 \\
    \midrule 
        CM-mix  & 77.0 & 77.9 & 52.7 \\
        CM-mix + Hard negatives & 77.5 & 78.3 & 51.5 \\
        CM-mix + OA Task 1 & 74.3 & 75.2 & 46.7 \\
        \midrule
        OA Tasks 2, 3, 4 & 75.3 & 75.7 & 47.5 \\
        CM-mix + OA Tasks 2, 3, 4 & 79.9 & 78.6 & 55.8  \\
        \midrule
        CM-mix + OA-mix & 80.1 ({\color{ForestGreen} +9.4}) & 81.1 ({\color{ForestGreen} +8.7}) & \textbf{58.9} ({\color{ForestGreen} +8.5}) \\
        CM-mix + Hard negatives + OA-mix & \textbf{80.4} ({\color{ForestGreen} +9.7}) & \textbf{81.8} ({\color{ForestGreen} +9.4}) & 58.6 ({\color{ForestGreen} +8.2}) \\
    \bottomrule
    \end{tabular}

    \caption{Comparisons of different task mixtures. Our Cross-Modal mix, CM-mix (using all four CM tasks), provides strong pre-training. Mixing all Cross-Modal and Object-Aware tasks to train together, achieves best results. The results also show the benefit of adding Object-Aware language tasks, and also the importance of the tasks themselves (e.g., adding only the hardest of them alone is not beneficial). For reference, as baselines, we show tasks that are commonly used in prior work: Captioning and MLM approaches. Large gains over the baselines are observed. 
    Multiple VQA datasets are used for downstream evaluation, which are trained and evaluated with open-vocabulary.}
    \label{tab:oi-mix}
\end{table*}

\begin{table*} 
    \begin{subtable}{.25\linewidth}
      \caption{Task 2: Object Exists.}
      \centering
    \begin{tabular}{l|cc}
    \toprule
         & Easy & Hard \\
    \midrule
        GQA & \textbf{71.6} & 71.3 \\
        SNLI-VE & \textbf{69.5} & 67.5 \\
        VQA & 47.8 & \textbf{48.6} \\
    \bottomrule
    \end{tabular}
    \end{subtable}\hfill%
    \begin{subtable}{.3\linewidth}
      \centering
        \caption{Task 3: And/Or Objects.}
    \begin{tabular}{l|cc}
    \toprule
         & Easy & Hard \\
    \midrule
        GQA & 70.6 & \textbf{71.5} \\
        SNLI-VE & \textbf{74.9} & 73.3 \\
        VQA & 42.6 & \textbf{45.3} \\
    \bottomrule
    \end{tabular}
    \end{subtable}\hfill%
    \begin{subtable}{.3\linewidth}
      \centering
        \caption{Task 4: Which Objects.}
    \begin{tabular}{l|cc}
    \toprule
         & Easy & Hard \\
    \midrule
        GQA & \textbf{74.6} & 74.0 \\
        SNLI-VE & 76.1 & \textbf{76.7} \\
        VQA & 44.6 & \textbf{46.8} \\
    \bottomrule
    \end{tabular}
    \end{subtable}
     \caption{`Easy' vs. `hard' negatives for the three object-aware tasks.}
    \label{tab:neg}
\end{table*}

\begin{figure}
    \centering
    \includegraphics[width=\linewidth]{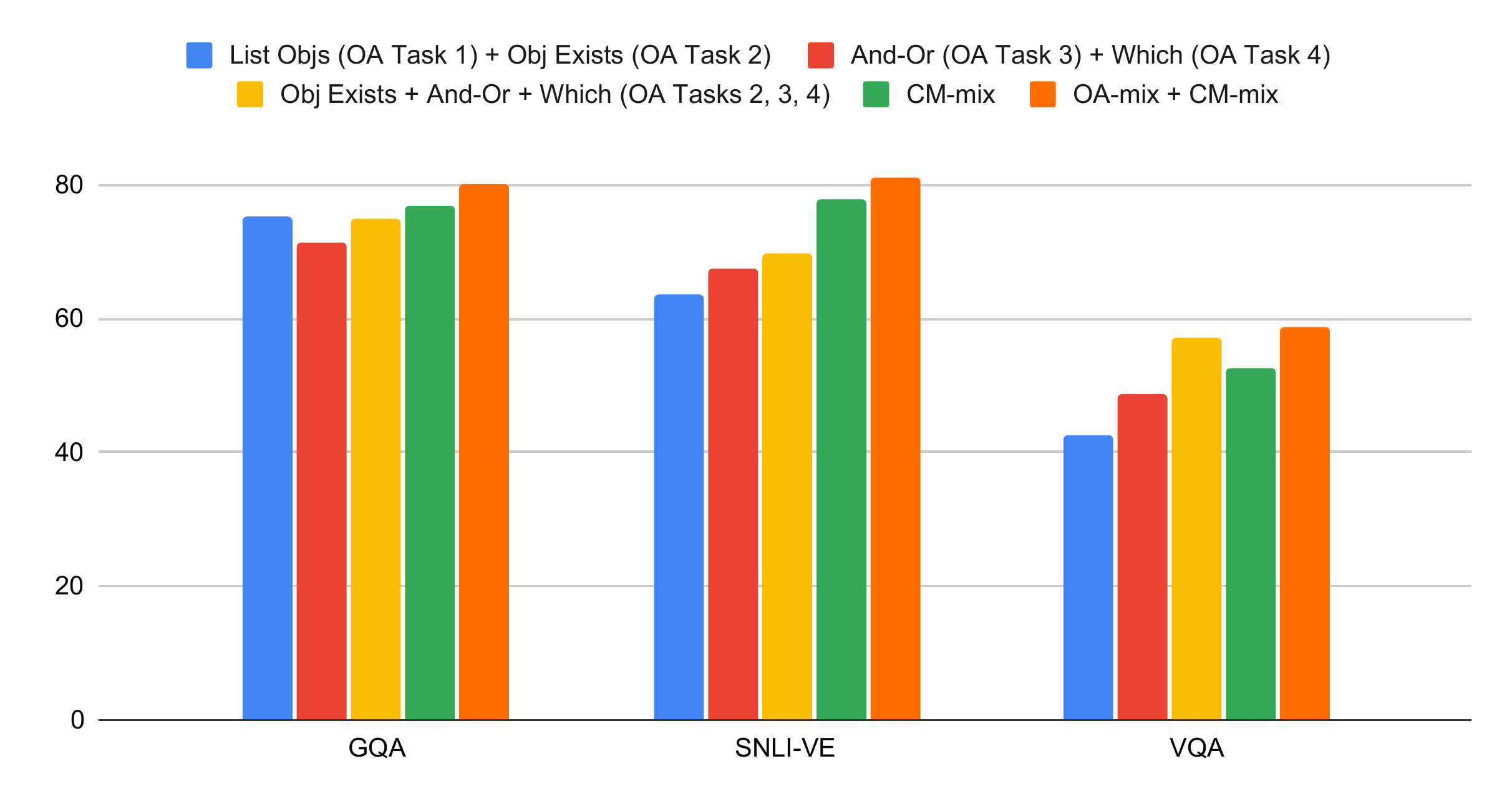}
    \caption{Mixture Results. Performance results for OA mixtures and CM mixtures. We find that a mixture of both the CM pre-training tasks and a mixture of OA pre-training tasks performs best across all downstream datasets, and outperforms both mixtures of CM-only and OA-only tasks. }
    \label{fig:mixture}
\end{figure}

We then evaluate combination of these OA tasks (Fig. \ref{fig:mixture}), and also compare with the performance of mix of CM tasks, which we observe to work very well (Fig.~\ref{fig:ln-pretrain}). Finally, we mix all of the OA and CM tasks, which we find to perform the best (Fig. \ref{fig:mixture}), even improving CM-mix and OA-mix. 

We further summarize the main results in Table \ref{tab:oi-mix}. We can see there the advantage of using the proposed mixture pre-training which uses both the CM-mix and OA-mix together. There are also large gains over the baseline methods, with improvements ranging from \textbf{+2.5\%} to \textbf{+9.7\%}. We note that our model is multi-task, of modest size, and the pre-training dataset is only limited to OpenImages and associated text, and that our results are generated in the open-vocabulary setting where only exact matches are counted as correct. Nevertheless, the model works well compared to SOTA trained on large datasets and with additional fine-tuning~\cite{tan2019lxmert,simvlm}, outperforming best SOTAs on GQA~\cite{tan2019lxmert,zhang2021vinvl}.
 Hard negatives 
 can additionally help. 

We then show more results and introspection into the model in Table \ref{tab:oi-mix}.
Specifically, we include results separating the Objects-Aware (OA) Task 1, from the other OA tasks.  For example, we add the `list all objects' task to the CM-mix pre-training. As seen, the task alone, when added to a CM-mixture, 
is actually detrimental to the CM-mix
(-3\% on average across the VQA datasets). This is interesting as all tasks in the mixture work really well. We hypothesize this is due to OpenImages not being exhaustively annotated. I.e., there are many objects in OpenImages that are not annotated. When the model is asked to list all the objects, if it lists an object that is visible, but is not annotated, it will be penalized in this task, even though the model is correct in the prediction. Task 1 is also much more challenging than others. 
We further observe that when mixing together the other three OA tasks `object exists', `and-or', and `which objects' with CM-mix, 
the overall performance is better. 
These tasks have the model learn similar things, without penalizing the model for non-annotated objects. This mixture improve over the single tasks by an average of 2\%
 (Table \ref{tab:oi-mix}).
When training all tasks together, including the challenging Task 1, we see that the task now is marginally helpful when included to the other three OA tasks, since now, in the context of others, it can help to correlate specific objects and semantics concepts between the text and the image. 

 
 
\begin{figure}
    \centering
    \includegraphics[width=0.8\linewidth]{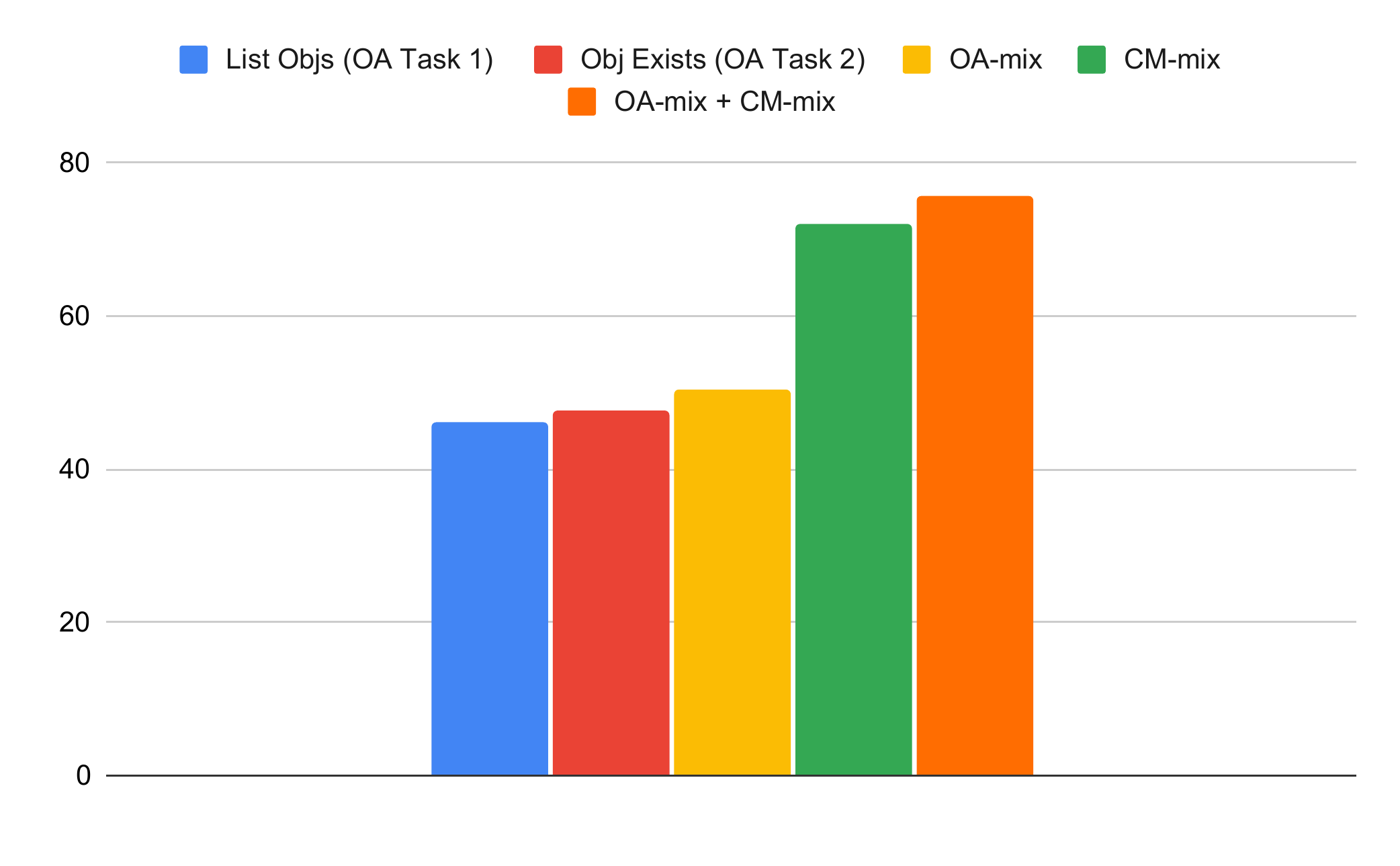}
    \caption{Evaluation of pre-training on an image captioning task (COCO captions~\cite{chen2015cococaptions}). CIDEr scores are shown. Similarly to the VQA case, mixing the pre-training tasks, both cross-modal image-text (CM) and object-aware (OA), achieves the best caption scores.}
    \label{fig:coco-scores}
\end{figure}

\subsection{Results on Image Captioning}
\label{sec:results_cap}

We then evaluate the pre-training above on an image captioning task~\cite{chen2015cococaptions}, shown in Fig. \ref{fig:coco-scores}. While these numbers are low compared to SOTA models, we focus here on the relative gains from the various tasks. Here, we see CM-mix works well, as it is a captioning task. However, combining  and OA mixtures, as above, further improves performance on this task, showing the benefit of adding object-aware data to the pre-training mix, even for captioning.

\subsection{Ablation experiments with hard negatives}

We compare the hard and easy negatives for the three object-aware tasks (Tasks 2, 3, and 4). The results are summarized in Table \ref{tab:neg}. We find that for Object-Aware tasks, there is not a drastic difference between the hard and easy negatives, though the hard negatives generally perform slightly better for most VQA tasks, especially the `which objects' task. This is an encouraging sign as many datasets do not have hard negative annotations, so adding them to the pre-training is another 
avenue for
improvements 
(Table~\ref{tab:oi-mix}).

\section{Conclusions}
We presented a pre-training approach to use image-language datasets to train image-text transformer models. We found that while captioning data is the strongest single-task pre-training, the addition of multiple 
object-aware tasks, improves the model performance across VQA, visual entailment and captioning.


{\small
\bibliographystyle{ieee_fullname}
\bibliography{egbib}

\begin{thebibliography}{10}\itemsep=-1pt

\bibitem{agrawal2015vqa}
Aishwarya Agrawal, Jiasen Lu, Stanislaw Antol, Margaret Mitchell, C.~Lawrence
  Zitnick, Dhruv Batra, and Devi Parikh.
\newblock Vqa: Visual question answering.
\newblock In {\em ICCV}, 2015.

\bibitem{chen2015cococaptions}
Xinlei Chen, Hao Fang, Tsung-Yi Lin, Ramakrishna Vedantam, Saurabh Gupta, Piotr
  Dollar, and C.~Lawrence Zitnick.
\newblock Microsoft coco captions: Data collection and evaluation server.
\newblock In {\em https://arxiv.org/abs/1504.00325}, 2015.

\bibitem{chen2020uniter}
Yen-Chun Chen, Linjie Li, Licheng Yu, Ahmed~El Kholy, Faisal Ahmed, Zhe Gan, Yu
  Cheng, and Jingjing Liu.
\newblock Uniter: Universal image-text representation learning.
\newblock In {\em ECCV}, 2020.

\bibitem{hudson2019gqa}
Drew~A Hudson and Christopher~D Manning.
\newblock Gqa: a new dataset for compositional question answering over
  realworld images.
\newblock In {\em CVPR}, 2019.

\bibitem{align}
Chao Jia, Yinfei Yang, Ye Xia, Yi-Ting Chen, Zarana Parekh, Hieu Pham, Quoc~V.
  Le, Yunhsuan Sung, Zhen Li, and Tom Duerig.
\newblock Scaling up visual and vision-language representation learning with
  noisy text supervision.
\newblock In {\em ICML}, 2021.

\bibitem{openimages}
Alina Kuznetsova, Hassan Rom, Neil Alldrin, Jasper Uijlings, Ivan Krasin, Jordi
  Pont-Tuset, Shahab Kamali, Stefan Popov, Matteo Malloci, Alexander
  Kolesnikov, et~al.
\newblock The open images dataset v4.
\newblock {\em ICCV}, 128(7):1956--1981, 2020.

\bibitem{li2020oscar}
Xiujun Li, Xi Yin, Chunyuan Li, Pengchuan Zhang, Xiaowei Hu, Lei Zhang, Lijuan
  Wang, Houdong Hu, Li Dong, Furu Wei, Yejin Choi, and Jianfeng Gao.
\newblock Oscar: Object-semantics aligned pre-training for vision-language
  tasks.
\newblock In {\em ECCV}, 2020.

\bibitem{vilbert2020}
Jiasen Lu, Dhruv Batra, Devi Parikh, and Stefan Lee.
\newblock Vilbert: Pretraining task-agnostic visiolinguistic representations
  for vision-and-language tasks.
\newblock In {\em CVPR}, 2019.

\bibitem{localizednarratves}
Jordi Pont-Tuset, Jasper Uijlings, Soravit Changpinyo, Radu Soricut, and
  Vittorio Ferrari.
\newblock Connecting vision and language with localized narratives.
\newblock In {\em ECCV}, 2020.

\bibitem{clip}
Alec Radford, Jong~Wook Kim, Chris Hallacy, Aditya Ramesh, Gabriel Goh,
  Sandhini Agarwal, Girish Sastry, Amanda Askell, Pamela Mishkin, Jack Clark,
  Gretchen Krueger, and Ilya Sutskever.
\newblock Learning transferable visual models from natural language
  supervision.
\newblock In {\em ICML}, 2021.

\bibitem{T5}
Colin Raffel, Noam Shazeer, Adam Roberts, Katherine Lee, Sharan Narang, Michael
  Matena, Yanqi Zhou, Wei Li, and Peter~J. Liu.
\newblock Exploring the limits of transfer learning with a unified text-to-text
  transformer.
\newblock In {\em Journal of Machine Learning Research}, 2020.

\bibitem{cc3m}
Piyush Sharma, Nan Ding, Sebastian Goodman, and Radu Soricut.
\newblock Conceptual captions: A cleaned, hypernymed, image alt-text dataset
  for automatic image captioning.
\newblock In {\em ACL}, 2018.

\bibitem{tan2019lxmert}
Hao Tan and Mohit Bansal.
\newblock Lxmert: Learning cross-modality encoder representations from
  transformers.
\newblock In {\em EMNLP}, 2019.

\bibitem{xiao2021early}
Eric Mintun Trevor Darrell Piotr Dollár Ross~Girshick Tete~Xiao, Mannat~Singh.
\newblock Early convolutions help transformers see better.
\newblock In {\em NeurIPS}, 2021.

\bibitem{vaswani2017attention}
Ashish Vaswani, Noam Shazeer, Niki Parmar, Jakob Uszkoreit, Llion Jones,
  Aidan~N. Gomez, Lukasz Kaiser, and Illia Polosukhin.
\newblock Attention is all you need.
\newblock In {\em NeurIPS}, 2017.

\bibitem{wang2022objectaware}
Alex~Jinpeng Wang, Yixiao Ge, Guanyu Cai, Rui Yan, Xudong Lin, Ying Shan,
  Xiaohu Qie, and Mike~Zheng Shou.
\newblock Object-aware video-language pre-training for retrieval.
\newblock In {\em CVPR}, 2022.

\bibitem{simvlm}
Zirui Wang, Jiahui Yu, Adams~Wei Yu, Zihang Dai, Yulia Tsvetkov, and Yuan Cao.
\newblock Simvlm: Simple visual language model pretraining with weak
  supervision.
\newblock In {\em https://arxiv.org/pdf/2108.10904.pdf}, 2021.

\bibitem{snli-ve}
Ning Xie, Farley Lai, Derek Doran, and Asim Kadav.
\newblock Visual entailment: A novel task for fine-grained image understanding.
\newblock In {\em https://arxiv.org/abs/1901.06706}, 2019.

\bibitem{zhang2021vinvl}
Pengchuan Zhang, Xiujun Li, Xiaowei Hu, Jianwei Yang, Lei Zhang, Lijuan Wang,
  Yejin Choi, and Jianfeng Gao.
\newblock Vinvl: Revisiting visual representations in vision-language models.
\newblock In {\em Proceedings of the IEEE/CVF Conference on Computer Vision and
  Pattern Recognition}, pages 5579--5588, 2021.

\end{thebibliography}
}

\end{document}